\documentclass[journal]{IEEEtran}
\IEEEoverridecommandlockouts  
\usepackage{amsmath}
\usepackage{algorithm}
\usepackage[noend]{algpseudocode}
\usepackage{amsfonts}
\usepackage[subrefformat=parens,labelformat=parens]{subfig}
\usepackage{graphics} 
\usepackage{graphicx}
\usepackage{amssymb}
\usepackage{caption}
\usepackage{array}
\usepackage{amsmath}
\usepackage{hyperref}
\usepackage{mathtools}
\usepackage{varwidth}
\usepackage{multicol}
\usepackage{pgfplots}
\usepackage{cite}
\usepackage{tikz}
\usetikzlibrary{pgfplots.units}
\hypersetup{bookmarks=false, hidelinks=true}
\begin{document}

\title{VDB-EDT: An Efficient Euclidean Distance Transform Algorithm Based on VDB Data Structure}
\author{Delong Zhu$^1$,~\IEEEmembership{Student Member,~IEEE,} Chaoqun Wang$^1$,~\IEEEmembership{Member,~IEEE,} Wenshan Wang$^2$,~\IEEEmembership{ Member,~IEEE,} \\Rohit Garg$^{2}$,~\IEEEmembership{Member,~IEEE,} Sebastian Scherer$^{2*}$,~\IEEEmembership{Senior Member,~IEEE,} Max Q.-H. Meng$^{1*}$,~\IEEEmembership{Fellow,~IEEE}
\thanks{$^{1}$ The authors are with the Department of Electronic Engineering, The Chinese University of Hong Kong, Shatin, N.T., Hong Kong SAR, China. \textit{email: \{dlzhu, cqwang, qhmeng\}@ee.cuhk.edu.hk}
}
\thanks{$^{2}$ The authors are with the Robotics Institute, Carnegie Mellon University, USA. \textit{email: \{wenshanw, rg1, basti\}@andrew.cmu.edu}
}
\thanks{$*$ The corresponding authors, and this project is partially supported by the Hong Kong RGC GRF grants \#14200618 and Hong Kong ITC ITSP Tier 2 grant \#ITS/105/18FP awarded to Max Q.-H. Meng. }
}

\maketitle
\begin{abstract}
This paper presents a fundamental algorithm, called VDB-EDT, for Euclidean distance transform (EDT) based on the VDB data structure. 
The algorithm executes on grid maps and generates the corresponding distance field for recording distance information against obstacles, which forms the basis of numerous motion planning algorithms.
The contributions of this work mainly lie in three folds. 
Firstly, we propose a novel algorithm that can facilitate distance transform procedures by optimizing the scheduling priorities of transform functions, which significantly improves the running speed of conventional EDT algorithms.
Secondly, we for the first time introduce the memory-efficient VDB data structure, a customed B+ tree, to represent the distance field hierarchically.  
Benefiting from the special index and caching mechanism, VDB shows a fast (average \textit{O}(1)) random access speed, and thus is very suitable for the frequent neighbor-searching operations in EDT.  
Moreover, regarding the small scale of existing datasets, we release a large-scale dataset captured from subterranean environments to benchmark EDT algorithms. Extensive experiments on the released dataset and publicly available datasets show that VDB-EDT can reduce memory consumption by about 30$\%$-85$\%$, depending on the sparsity of the environment, while maintaining a competitive running speed with the fastest array-based implementation. The experiments also show that VDB-EDT can significantly outperform the state-of-the-art EDT algorithm in both runtime and memory efficiency, which strongly demonstrates the advantages of our proposed method. The released dataset and source code are available on \url{https://github.com/zhudelong/VDB-EDT}.
\end{abstract}



\section{Introduction}
\label{introduction}
\begin{figure}[t]
	\centering
	\includegraphics[width=0.45\textwidth]{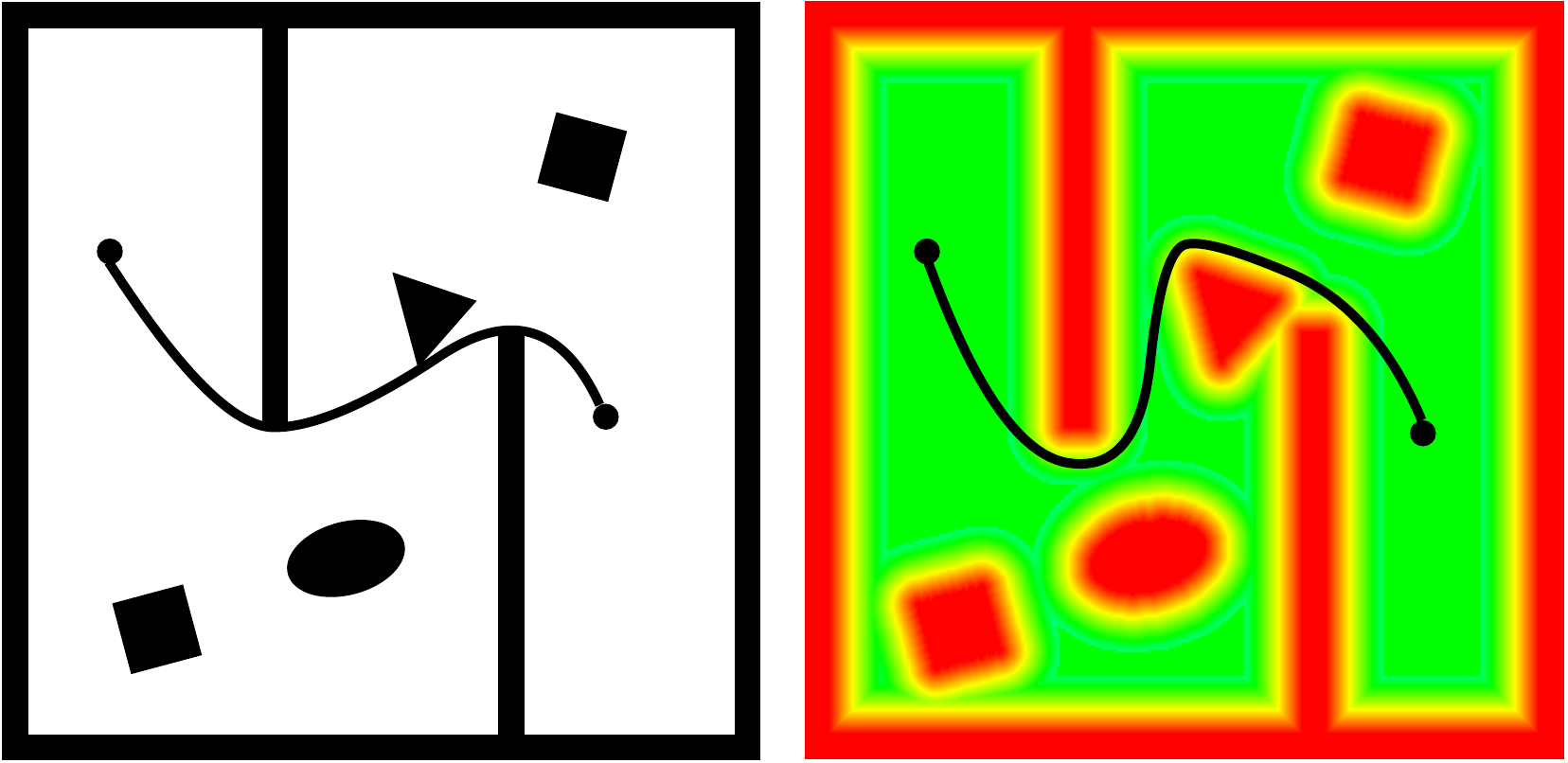}
	\caption{Demonstration of motion planning based on the distance field. The left figure shows a grid map, in which the trajectory is planned based on a path length cost. The right figure visualizes the generated distance field, and the trajectory is pushed away from obstacles by additionally minimizing a clearance cost defined on the distance field.} 	
	\label{staticplan}
\end{figure}

EDT algorithm plays a fundamental role in robotic motion planning problem. The generated distance field, which provides distance information against obstacles, is an indispensable ingredient for planning safe trajectories \cite{kalakrishnan2011stomp, ZuckerCHOMP, 8876884, Oleynikova2016Continuous, 7932539, usenko2017real}. As demonstrated in Fig. \ref{staticplan}, by additionally minimizing a clearance cost defined on the distance field, the optimal trajectory is pushed away from obstacles.
EDT can be performed globally on static grid maps \cite{felzenszwalb2012distance}, referred to as \textit{global transform}, like the one in Fig. \ref{staticplan}. It can also incrementally execute to tackle dynamic changes in the environment, referred to as \textit{incremental transform}, which is particularly useful for applications that rely on online trajectory generation, e.g., robotic explorations \cite{zhu2017hawkeye, zhu2018deep, tingguang2018exp, chaoqun2019Semantic}, Micro Aerial Vehicle (MAV) based inspections and transportations \cite{Wallar2015, leetrans2018, bonatti2018autonomous, kangcheng2019}. 
However, EDT is a computationally expensive algorithm. The building and maintenance of distance fields usually cost a large number of computing resources, which is considered a bottleneck for numerous motion planning algorithms.
In this paper, we focus on improving the efficiency of EDT algorithms by optimizing the fundamental data structure and distance transform procedures. 

There are many well-developed EDTs \cite{kalra2009incremental, scherer2009efficient, lau2010improved, scherer2012river, cover2013sparse, Oleynikova_vox, vespa2018efficient, FIESTA} for building distance fields, but due to the fact that the runtime efficiency and memory efficiency cannot be achieved simultaneously, there does not exist a dominant approach in real-world applications. In conventional implementations, the multi-dimensional array is a commonly-used data structure, as it is easy to implement and with the fastest random access speed.
Drawbacks of these implementations are that the memory for maintaining distance fields should be fully allocated before launching the transform process, and thus the field size must be known in advance, which makes the methods not suitable to transform dynamically-growing maps in applications like robotic explorations.
Moreover, the full allocation of memory yields a great waste, because in grid maps, there exist a large number of unreachable or uninterested regions, and it is typically unnecessary to allocate memory for transforming these regions. 
To overcome these drawbacks, the hashing-based EDTs are proposed in recent studies \cite{Oleynikova_vox, FIESTA}, in which the memory is dynamically allocated based on the \textit{voxel hashing} technique \cite{niessner2013real}. Although the random access speed is sacrificed to some extent, the hashing-based methods significantly reduce the memory consumption of EDTs, enabling their successful deployment in MAV platforms for online trajectory generation.   

In this paper, we further extend the idea of hashing-based EDTs by leveraging tree structures to represent distance fields hierarchically, referred to as \textit{hierarchical hashing}. This idea is based on our observations of the planning process, in which the optimal trajectory typically lies within a certain range around obstacles (see Fig. \ref{staticplan}), and the full distance information is redundant for the planning process. 
Therefore, we propose to limit the transform within a certain distance range, and take the maximum range as the distance value for untransformed map regions 
(the green area in Fig. \ref{staticplan}). These regions can then be efficiently encoded by a few number of tree nodes. Such a property is called \textit{spatial coherency}, which can help further reduce memory consumption. 
For the tree structure, we introduce the novel VDB \cite{museth2013vdb}, a volumetric and dynamic B+ tree, rather than the widely-adopted Octree \cite{meagher1982geometric, hornung2013octomap, vespa2018efficient}, as the underlying implementation.
Benefiting from the fast index and caching systems, VDB achieves a much faster random access speed than Octree and also exhibits a competitive performance with the voxel hashing. To the best of our knowledge, this is the first work that adopts VDB for distance field representation in the robotic area.       

Another key contribution of this work lies in the development of a novel transform procedure. Typically, an EDT algorithm includes two functions for updating distance fields: a \textit{raise} function used for clearing the regions affected by removed obstacles, and a \textit{lower} function used for updating the regions affected by newly added obstacles. The two functions are scheduled according to a certain priority to maintain a consistent distance field.
Efficient function implementations and well-developed scheduling strategies are crucial for accelerating the transform procedure, which is also the focus of EDT algorithms \cite{scherer2009efficient, lau2010improved, scherer2012river, cover2013sparse}. 
Our improvement mainly lies in optimizing the scheduling priority of the \textit{raise} and \textit{lower} functions by additionally introducing a flag variable to record the transform status for each map cell, according to which, the affected regions are precisely identified and processed by the functions, thus avoiding a lot of redundant operations.

The contributions of this paper are summarized as follows:  
\begin{itemize}
	\item We for the first time introduce the VDB data structure for distance field representation, which significantly reduces the memory consumption of EDT.
	\item We propose a novel algorithm to facilitate distance transform procedure and significantly improve the running speed of conventional EDT algorithms.
	\item We release a large-scale dataset captured from a 200m-by-150m subterranean environment to benchmark EDT algorithms. The source code is also released as a ros-package to benefit the community.
	\item We conduct extensive experiments on simulated and real-world datasets, demonstrating the significant advantages and state-of-the-art performance of our method.    
\end{itemize}

The remainder is organized as follows. Section \ref{relatedwork} briefly surveys recent studies on EDTs. Section \ref{problem} and \ref{meth} detail the novel data structure and transform procedure, respectively. Experiment results and discussions are repented in Section \ref{secExp}. The paper is concluded in Section \ref{conclusion}.
\vspace{-5mm}
\begin{figure*}[t]
	\captionsetup{belowskip=-6pt}
	\centering
	\includegraphics[width=0.95\linewidth]{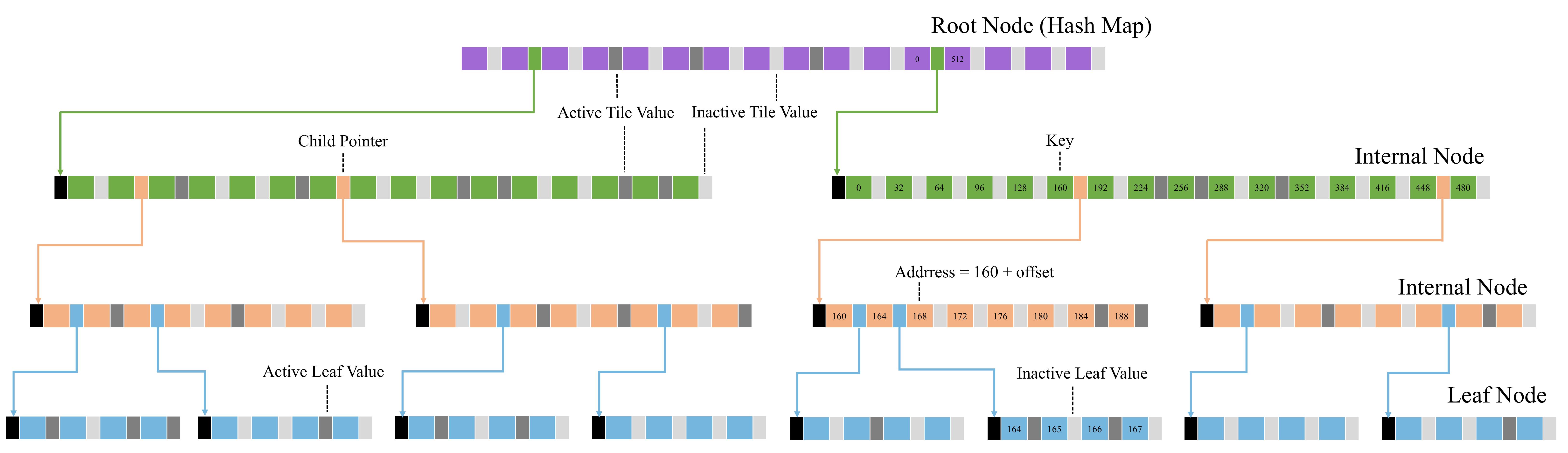}
	\caption{The 1-D structure of VDB. The top array represents a hash map of root nodes and the others are direct access tables storing internal or leaf nodes. Each node has a unique address and a data field. The address of the root node is calculated by a very efficient hash function. Internal nodes and leaf nodes are visited by direct access. The data field may store a child pointer or a tile value, which is identified by a flag variable. Each data field is also associated with an active flag, which can be used to indicate node state.    }
	\label{arch}
\end{figure*} 
\section{Related Work}
\label{relatedwork}
EDTs can be divided into two categories according to whether the algorithm depends on specialized data structures.

\textbf{General EDTs:} The general EDT can be implemented with multi-types data structures. One of the most influential EDT algorithms is the \textit{Brushfire} \cite{brush1991}, which is developed based on the \textit{Dijkstra}'s algorithm, and can only perform global transform. Once the environment changes, it has to update the entire distance field.    
To make the algorithm cope with dynamic environments, Nidhi \textit{et al.} \cite{kalra2009incremental} propose a variant of the Brushfire based on $D^*$ algorithms \cite{stentz1997optimal, koenig2002d, 7878681}, which forms the first incremental EDT in the literature. The new algorithm only updates the affected regions in the distance field, thus avoiding a large number of unnecessary calculations. 
Based on the incremental Brushfire, Scherer \textit{et al.} \cite{scherer2009efficient} present a Limited Incremental Distance Transform (LIDT) algorithm, in which the authors set a distance threshold to limit the transform scope, and regions out of the scope are no longer considered. The LIDT has a similar transform procedure with \cite{kalra2009incremental} but achieves a better running efficiency benefiting from the reduction of transform scope. Based on \cite{kalra2009incremental} and \cite{scherer2009efficient}, Lau \textit{et al.} \cite{lau2010improved} further optimize the transform procedure by removing redundant map queries, and present a more compact EDT implementation, which avoids a significant number of transform steps and achieves very high efficiency.
Afterward, Scherer \textit{et al.} \cite{scherer2012river} and Cover \textit{et al.} \cite{cover2013sparse} propose an even more efficient implementation of the \textit{lower} function, achieving high-frequency motion planning process. In this work, we further push the limits of general EDT algorithms with an efficient scheduling strategy of the \textit{lower} and \textit{raise} functions. 
     
\textbf{Specialized EDTs:} In recent studies, researchers focus on designing special data structures and customizing transform procedures to improve EDTs.
Oleynikova \textit{et al.} \cite{Oleynikova_vox} propose the Voxblox based on voxel hashing to incrementally build Euclidean Signed Distance Field (ESDF) from the Truncated Signed Distance Field (TSDF), which has a faster mapping speed than the commonly-used occupancy grid map. By sufficiently leveraging the distance information in TSDF, Voxblox achieves a real-time performance on single computing thread, and allows dynamically-growing map size.   
Emanuele \textit{et al.} \cite{vespa2018efficient} propose a volumetric-SLAM framework for TSDF and occupancy mapping, in which the voxel hashing is combined with Octrees as a hierarchical data structure to represent the constructed maps. This work shares some similarities with ours in terms of the hierarchical map representation. The presented data structure is also compatible with our proposed algorithm, but due to the slow query speed of Octree, it is not adopted in this paper. 
Han \textit{et al.} \cite{FIESTA} demonstrate a similar idea with \cite{vespa2018efficient}, and propose the FIESTA framework to combine voxel hashing with doubly linked list to represent distance field, in which the list is used to group field points that share the same obstacle. When the obstacles are removed, the affected field regions can be efficiently visited by simply iterating the lists, thus avoiding the graph-searching process in general EDTs. FIESTA can significantly outperform Voxblox in runtime efficiency, and as the state-of-the-art method, it is adopted as one of the baselines in this work.

\section{Problem Definition}
\label{problem}
A typical distance transform problem on a grid map $\mathcal{M}$ can be expressed as follows:
\begin{equation}
\label{edt}
\begin{aligned}
d&(x_i) = \min_{x_j} f(x_i, x_j), \\
\textrm{s.t.}  \quad  & x_i \in \mathcal{M}_f, x_j \in \mathcal{M}_o,\\
\end{aligned}
\end{equation}
where $x_i$ and $x_j$ represent the coordinates of map cells in $\mathcal{M}$, and $\mathcal{M}_f, \mathcal{M}_o \subset \mathcal{M}$ are the sets of free cells and occupied cells (obstacles), respectively. The objective function $f(x_i, x_j)$ measures the distance between $x_i$ and $x_j$. The squared Euclidean distance and Manhattan distance are commonly-used measurements in most planning scenarios. Eq. \eqref{edt} indicates that the goal of distance transform is to find out the closest occupied cell $x_j^*$ and calculate the distance $f(x_i, x_j^*)$ for each free cell $x_i$. EDT is essentially a search-based optimization framework for solving the problem defined in Eq. (\ref{edt}).

The output $d(x)$ of EDT, i.e., the distance field, is used to define the clearance cost for generating safe trajectories, which can be achieved by solving the following optimization problem:   
\begin{equation}
\label{plan}
\begin{aligned}
\min_{x_{0:N}} \quad & \sum_{i=0}^{N} \alpha||x_{i+1}-x_i|| + (1-\alpha) \text{max}(0, d_{max}-d(x_i)),\\
\textrm{s.t.}  \quad & \quad\quad\quad x_i,x_{i+1} \in \mathcal{M}_f,\\
					 & \quad\quad\quad x_0=x_s, x_N=x_f, \\
					 & \quad\quad\quad{g(x_i, x_{i-1}, x_{i+1}) < \theta},   \\
\end{aligned}
\end{equation}
where $d_{max}$ is the maximum transform distance, $x_s$ is the start cell, $x_f$ is the goal cell, $\alpha$ is the balancing coefficient, and $g < \theta$ is used to constrain the angle between consecutive segments to smooth the trajectory. The first and second terms in objective function are path length cost and clearance cost, respectively. Herein, as the clearance cost indicated, field points with a distance value of $d(x_i) \geq d_{max}$ have no influence on the optimization process. We thus only need to perform a limited distance transform within the map regions that meet the constraint of $ f(x_i, x_j) < d_{max}$. Under this constraint, Eq. \eqref{edt} becomes a LIDT problem \cite{scherer2009efficient}.


\begin{algorithm*}[t]
	\caption{VDB Based Incremental Euclidean Distance Transform}
	\label{alg}
	\algrenewcommand\textproc{}
	\algrenewcommand\algorithmicprocedure{}
	\algrenewcommand\algorithmicthen{}
	\algrenewcommand\algorithmicrequire{DDD}
	\algrenewcommand\algorithmicindent{1.0em}%
	\vspace{-0.4cm}
	\begin{multicols}{3}
		\begin{algorithmic}[1]
			\Statex{\textbf{Initialize}($s$)}
			\begin{algorithmic}[1]
				\State $s.obst \gets \emptyset$ 
				\State $s.dist \gets d_{max}$   
				\State $s.raise \gets notRaise$  
				\State $s.state \gets notQueued$ 
				\State m.setBackground($s$)
				\algstore{myalg}
			\end{algorithmic}
		\end{algorithmic}
		\vspace{0.1cm}	
		\begin{algorithmic}[1]
			\Statex{\textbf{SetObstacle}($s$)}
			\begin{algorithmic}[1]
				\algrestore{myalg}
				\State $occ \gets$ m.isObst($s$)
				\State $state \gets$ m.isQueued($s$)
				\If{$(\lnot\; occ) \wedge (\lnot\; state)$}
				\State $s.obst \gets s$
				\State $s.dist \gets 0$
				\State \underline{$s.raise \gets notRaise$}  \Comment -1
				\State $s.state \gets Queued$
				\State q.insert($s$, $0$)   \Comment zero priority
				\State m.updateNode($s$)
				\EndIf 
				\algstore{myalg}
			\end{algorithmic}
		\end{algorithmic}
		\vspace{0.1cm}
		\begin{algorithmic}[1]
			\Statex{\textbf{RemoveObstacle}($s$)} 
			\begin{algorithmic}[1]
				\algrestore{myalg}
				\State $occ \gets$ m.isObst($s$)
				\State $state \gets$ m.isQueued($s$) 
				\If{$occ \wedge (\lnot\; state)$}
				\State $s.obst \gets \emptyset$
				\State $s.dist \gets d_{max}$
				\State \underline{$s.raise \gets 0$} \Comment start to raise
				\State $s.state \gets Queued$
				\State q.insert($s$, $0$) 
				\State m.updateNode($s$)
				\EndIf
				\algstore{myalg}
			\end{algorithmic}  
		\end{algorithmic}
		\newpage
		\begin{algorithmic}[1]
			\Statex{\textbf{DistanceTransform}()}
			\begin{algorithmic}[1]
				\algrestore{myalg}
				\While{$\lnot\,$q.isEmpty()} 
				\State $s \gets$ q.pop()
				\If{$\lnot\,$m.isQueued($s$)}
				\State continue
				\EndIf
				\If{$s.raise \geq 0$}
				\State \Call{Raise}{$s$}
				\Else
				\State \Call{Lower}{$s$}
				\EndIf
				\EndWhile
				\algstore{myalg}
			\end{algorithmic}		
		\end{algorithmic}
		\begin{algorithmic}[1]
			\Statex{\textbf{Raise}($s$)}
			\begin{algorithmic}[1]
				\algrestore{myalg}
				\ForAll{$n \in $ Adj$_{26}$($s$)}
				\If{$n.obst = \emptyset$}
				\State continue 
				\EndIf
				\If{$\lnot\,$m.isObst($n.obst$)}
				\State \underline{$n.raise \gets n.dist$} \Comment key-1
				\State $n.state \gets Queued$
				\State q.insert($n$, $n.dist$) 
				\State $n.obst \gets \emptyset$
				\State $n.dist \gets d_{max}$
				\State m.updateNode($n$)
				\ElsIf{$n.state \neq Queued$}			  					 
				\State $n.state \gets Queued$
				\State q.insert($n$, $n.dist$)
				\State m.updateNode($n$)					  					  
				\EndIf
				\EndFor
				\State $s.raise \gets notRaise$
				\State $s.state \gets notQueued$
				\State m.updateNode($s$)
				\algstore{myalg}
			\end{algorithmic}		
		\end{algorithmic}
		\newpage	        
		\begin{algorithmic}[1]
			\Statex{\textbf{Lower}($s$)}
			\begin{algorithmic}[1]
				\algrestore{myalg}
				\ForAll{$n \in $ Adj$_{26}$($s$)}
				\State $d_{new} \gets$ sqDist($s.obst$, $n$)	
				\State $d_{new} \gets$ min($d_{new}, d_{max}$)
				\vspace{0.05cm}
				\If{\underline{$n.raise \geq d_{new}$}}   \Comment {key-2}
				\State $n.obst \gets s.obst$
				\State $n.dist \gets d_{new}$
				\State $n.raise \gets notRaise$
				\State $n.state \gets Queued$
				\State q.insert($n$, $d_{new}$)
				\State m.updateNode($n$)
				\State continue
				\vspace{0.05cm}
				\ElsIf{$n.raise < 0$}
				\State $occ \gets$ m.isObst($n.obst$)
				\State $equ \gets$ ($d_{new} = n.dist$)
				\State $less \gets$ ($d_{new} < n.dist$)					
				\If{$less \vee (equ \wedge \lnot\,occ)$}
				\State $n.obst \gets s.obst$
				\State $n.dist \gets d_{new}$
				\State $n.raise \gets notRaise$
				\If{$(d_{new} < d_{max})$}
				\State $n.state \gets Queued$
				\State q.insert($n$, $d_{new}$)
				
				\EndIf
				\State m.updateNode($n$)	
				\EndIf 									
				\EndIf	
				\EndFor
				\vspace{0.05cm}
				\State $s.raise \gets notRaise$
				\State $s.state \gets notQueued$
				\State m.updateNode($s$)
			\end{algorithmic}
		\end{algorithmic}	
	\end{multicols}	
	\vspace{-0.3cm}
\end{algorithm*}

\section{Methodologies}
\label{meth}
In this section, we first introduce the VDB data structure, and then detail the proposed distance transform procedures.	

\subsection{Data Structure}
\label{datastructure}
VDB data structure is originally proposed by Museth \cite{museth2013vdb} to represent large, sparse, and animated volumetric data. It sufficiently exploits the sparsity of volumetric data, and employs a variant of B+ tree \cite{beyer1972organization} to represent the data hierarchically. 
The efficient memory management and fast (average \textit{O}(1)) random access speed of VDB make it very suitable for large-scale distance transform problems. 

To demonstrate the working mechanism, we visualize the 1-D structure of VDB in Fig. \ref{arch}. As is indicated, VDB shares some favorable properties with a standard B+ tree. Especially, the branching factors are very large and variable, making the tree shallow and wide, which consequently increases the capacity of VDB, meanwhile reducing the query length from root nodes to leaves. Octree-based structures \cite{hornung2013octomap, vespa2018efficient} adopt an opposite design, i.e., deep and narrow, thus not fast enough for distance transform. 
VDB also has an essential difference with the B+ tree, i.e., it encodes values in internal nodes, called \textit{tile value}. 
The tile value and \textit{child pointer} exclusively use the same memory unit (see Fig. \ref{arch}), and a flag is additionally leveraged to identify the different cases. A tile value only takes up tens of bits memory but can represent a large area in the distance field (see the analysis about \textit{spatial coherency} in Section \ref{introduction}), which is the key feature leveraged to improve memory efficiency.
For instance, in LIDT-like algorithms \cite{scherer2009efficient, lau2010improved, scherer2012river, cover2013sparse}, we can initialize the VDB-based distance field with a distance threshold $d_{max}$, and then dynamically allocate memory, i.e., grow the tree, for transformed map cells. In this way, the memory consumption can be significantly reduced.

In addition to efficient memory management, VDB also has a special coordinate-based index system, which enables a fast random access speed.
As illustrated in Fig. \ref{arch}, when performing query operations, VDB first generates a hash key using the coordinate to locate the root node. If the child pointer of this root node is empty, a tile value will be returned; otherwise, an offset will be calculated to locate the child node. This process continues until a leaf node is reached or a tile value is returned. All the calculations during this process are based on bit-wise logic operations, thus can be efficiently finished. 
It can be seen that the query process in VDB is very different from the standard key-searching process in B+ tree. Based on the special index system, the query time of VDB is reduced from logarithmic complexity to constant-time complexity. 
Moreover, VDB adopts a caching system that can record recently visited nodes. Queries started from these nodes have an even shorter path than a complete root-to-leaf query, thus further improving the average random access speed. The core of EDTs is a series of neighbor searching operations, and it is extremely suitable to apply the caching system. 

\subsection{VDB-EDT Algorithm}
\label{methodalg}
VDB-EDT aims to solve the problem defined in Eq. \eqref{edt}. Details of the transform procedures are presented in Alg. \ref{alg}. In conventional EDTs, the \textit{Raise} function has a higher priority and cannot be interrupted by the \textit{Lower} function, which yields a large number of repetitive queries. In our method, we eliminate this constraint by leveraging an additionally defined variable, the \textit{transform status} (marked by underlines in Alg. \ref{alg}), to optimize the priorities of the two functions. It is worth noting that the presented transform procedures are also compatible with other data structures.

The distance field represented by VDB is essentially a sparse volumetric grid, and each field point is represented by a grid cell $s$ indexed by a 3-D coordinate. The grid cell is maintained in memory as a data record, which is composed of four designed data fields:  
  
\begin{itemize}
	\item $obst$: the coordinate of the indexed obstacle, which may be empty $\emptyset$, invalid, or valid, representing that no obstacle is indexed, the obstacle has been removed, and the obstacle is still valid, respectively. 	
	\item $dist$: the distance to the closest obstacle. If $dist \ge d_{max}$, $dist \gets d_{max}$, which is actually the radius of the later defined transform waves, and named \textit{transform radius}. 
	\item $raise$: the transform status of grid cell $s$, which can be $notRaise$, indicating \textit{not need to be raised}, or the transform radius of the later defined raising waves.
	\item $state$: a flag that marks current cell is queued or not. To accommodate with our new transform procedures, we allow the same cell to be queued multiple times.
\end{itemize}

To better clarify our method, several technical terms and functions are defined as follows:
 \begin{itemize}
 	\item \textit{raising wave}: when an obstacle is removed, the grid cells that index $s$ as the closest obstacle need to be sequentially reset, which generates a raising wave. 
 	\item \textit{lowering wave}: when an obstacle is added, its surrounding grid cells need to be checked, and corrected if necessary, which generates a lowering wave.
 	\item \textit{wave centers}: the grid cells that are newly occupied by obstacles or newly freed from obstacles.
 	\item \textit{wave fronts}: the cells that are going to be scheduled in the priority queue.
 	\item \textit{wave boundaries}: the cells that reach the maximum transform distance or at the boundaries between different transform waves.
 	\item \textit{m}: the distance field represented by VDB data structure.
 	\item \textit{q}: the priority queue used for scheduling transform waves.
 	\item \textit{isObst($s$)}: the short name for \textit{isObstacle($s$)}, used for validating whether $s$ is an obstacle.
 	\item \textit{sqDist($s_i$,$s_j$)}: the function used for calculating the squared Euclidean distance between inputs.
 	\item \textit{Adj$_{26}$($s$)}: the set of 26-connected neighbor cells of $s$.
 \end{itemize}

\textbf{Initialization Function}: In this function, all grid cells are set to default values (lines 1-5). Due to the special design of VDB, the distance field at this time only contains a root node storing the default values.
As the environment changes, obstacles are added or removed, and the grid cells are occupied or freed accordingly. To record these changes, two functions called \textit{SetObstacle} and \textit{RemoveObstacle} are provided, in which the changed cells are first updated and then registered to the VDB by a \textit{updateNode} function that implements the index and caching system.  
Specifically, if a grid cell is newly occupied, we zero its distance and index itself as the closest obstacle (lines 9-10); if a cell is newly freed, we clear its indexed obstacle and reset it to the default distance value (lines 18-19). These changed grid cells are wave centers used for generating transform waves to update the distance field. Since the lowering and raising waves have different transform procedures, we also need to distinguish their types (line 11 and 20), and then make them queued (lines 12-13 and 21-22) to initialize the scheduling process.

\textbf{DistanceTransform Function}: After seeding wave centers in the priority queue, distance transform is accomplished by this function, in which the \textit{lower} and \textit{raise} functions are scheduled according to their priorities.
As shown in Alg. \ref{alg}, each time the queue pops up a cell $s$, we first check whether it is previously processed (lines 26-27). If not, we then identify the wave type it belongs to. If it belongs to a raising wave, i.e., $s.raise \geq 0$, function \textit{Raise} will be called to clear the cells that index $s$ as the closest obstacle (lines 28-29); otherwise, i.e., $s.raise < 0$, function \textit{Lower} will be called to update cells that are newly cleared by raising waves or hold incorrect distance value (lines 30-31). 
Elements in the priority queue are sorted by their distance to the respective wave centers, i.e., transform radius, which means all the waves propagate at an equal speed. Transform waves start from wave centers, propagate according to transform radius, and stop at wave boundaries.

\textbf{Raise Function}: We process each of the neighbor cells to see if they can be developed as raising-wave fronts. The specific procedure is explained as follows. If a neighbor cell $n$ holds an empty value in $obst$ field, which means it has been processed by other raising waves or the \textit{initialization} function, we then ignore this cell and continue to process the next one (lines 33-34); otherwise, we further validate the occupancy of its indexed obstacle. 
If the indexed obstacle no longer exists, this neighbor cell will be identified as a raising-wave front. We then update its transform status and distance information to prepare it for next-round propagation (lines 35-41).    
It is worth noting that, different from conventional EDTs where the transform status is treated as a binary flag for indicating wave types, we update the transform status with the current transform radius of $n$ (see key-1 in Alg. \ref{alg}), which provides precise information for optimizing the scheduling process (as we shall see in next paragraph).     
If the indexed obstacle of the neighbor cell is valid, it means that the current raising wave comes to one of its wave boundaries, and if the neighbor cell is still not queued yet, we insert it into the queue (lines 42-45). The arrival of a raising wave at a boundary cell indicates its stop in that specific direction.    
After finishing processing all the neighbor cells, the current cell $s$ is completely reset to default values (lines 46-48). Finally, these cells will be gradually updated along with the propagation of lowering waves.

\begin{figure*}[t]
	\captionsetup{belowskip=-10pt}
	\centering
	\includegraphics[width=1.0\linewidth]{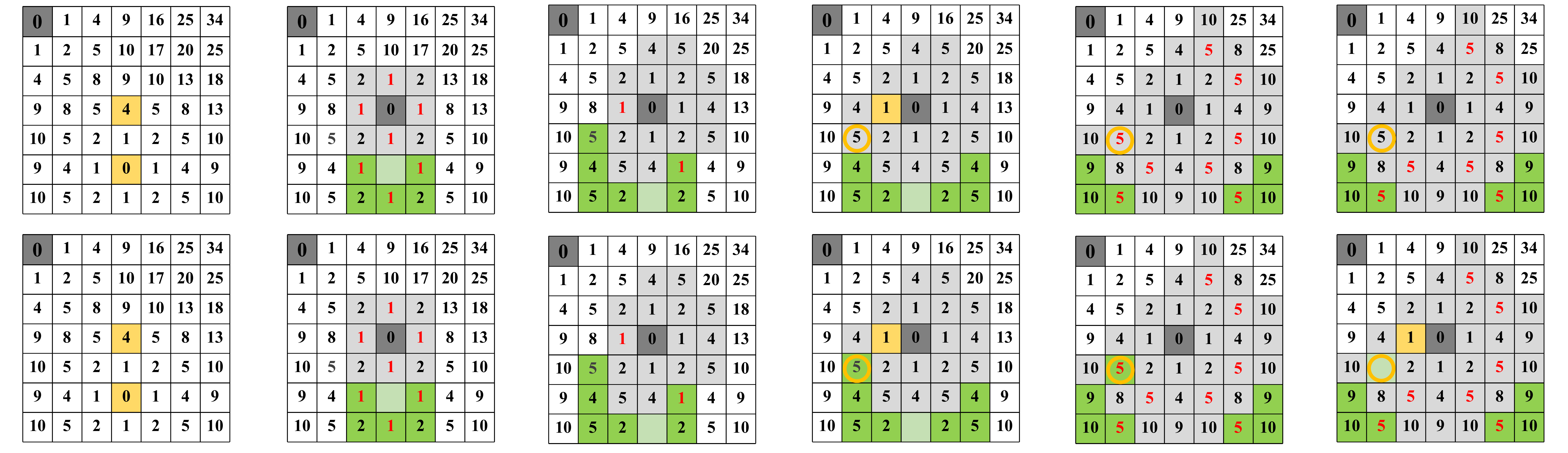}
	\caption{Transform procedures of our work (top row) and conventional EDTs (bottom row). The gray and green cells represent the lowering and raising wave, respectively. The cells indicated by red numbers are unprocessed wave fronts with the highest priority currently. The yellow color indicates the cell is under processing. The empty cells are the newly cleared ones, and the circles are used to highlight differences.}
	\label{procedure}
\end{figure*}

\textbf{Lower Function}: We check each of the neighbor cells and update the ones that can be lowered. As the procedure indicated, a truncated transform radius $d_{new}$ is first calculated for each neighbor cell (lines 50-51), and then by comparing $d_{new}$ with $n.raise$, we can identify the cells that can be lowered. Herein, three possible cases are enumerated:

\begin{itemize}
	\item $n.raise \geq d_{new}$: This indicates that the neighbor cell $n$ is a raising-wave front (since $n.raise \geq 0$), and it has already been queued (see lines 20-22 and 36-38) and well prepared for scheduling. Conventionally, we should not make any changes to this type of cell. However, since $n$ is more close to (or dominated by) the current wave center, it means that $n$ actually can be lowered. In this work, we propose to immediately lower $n$ and forcefully change it to a lowering-wave front (lines 52-59). Such a case cannot be identified or allowed by conventional EDTs, since the defined transform status, i.e., $raise$, in these methods is uninformative and cannot help make such decisions. 
		
	\item $n.raise < 0$: It means that the neighbor cell can be lowered, but its distance value may not be correct. Therefore, we have to further compare $d_{new}$ with $n.dist$ to identify the type of $n$: 1) If $d_{new}$ can strictly decrease $n.dist$, it means that $n$ is far away from its original wave center, and now is affected (or dominated) by the current wave; 2) If $n$ holds an equal $dist$ but indexes an invalid obstacle, it means that $n$ is an invalid wave boundary. The cell $n$ in both cases needs to be lowered, and be queued if $n$ does not exceed the maximum transform distance (lines 60-71), namely $n$ is developed as a lowering-wave front. 
	 
	\item $d_{new} > n.raise \geq 0$: It means that the current lowering wave meets a raising wave, but since $n$ is mainly affected (or dominated) by the raising wave, it cannot be lowered by the current lowering wave.       
\end{itemize}

After processing the neighbor cells, we reset the transform status of $s$, changing it from a lowering-wave front to a lowered cell (lines 72-74). 
Note that, in the first case, we improve the scheduling process by changing the type of wave fronts, which is the major difference between our method and conventional EDTs. In real-world applications, dynamic objects typically follow continuous trajectories, and thus the newly occupied and freed cells are located nearby, which provides more chances for the lowering waves to stop the raising waves at an early stage, as presented in the first case.

To better illustrate the procedures presented in Alg. \ref{alg}, we visualize an example in Fig. \ref{procedure}. The first column includes two distance fields, which are currently in the same state. We then add an obstacle to the yellow cell marked with 4 and remove an obstacle from the yellow cell marked with 0. After one-step propagation, the distance fields evolve to the second column, in which seven high-priority front cells need processing. The leftmost green cell denoted with 1 is first processed, and then another four wave fronts, yielding the states in the third column. The remaining ones are then processed and we get the fourth column, in which the differences start to appear (highlighted by yellow circles). It can be seen that the yellow cell denoted with 1 in the top row immediately lowers the raising-wave front and changes it into a lowering-wave front, which helps avoid at least $2*26$ queries (see the fifth and last columns) compared with the conventional method.
 
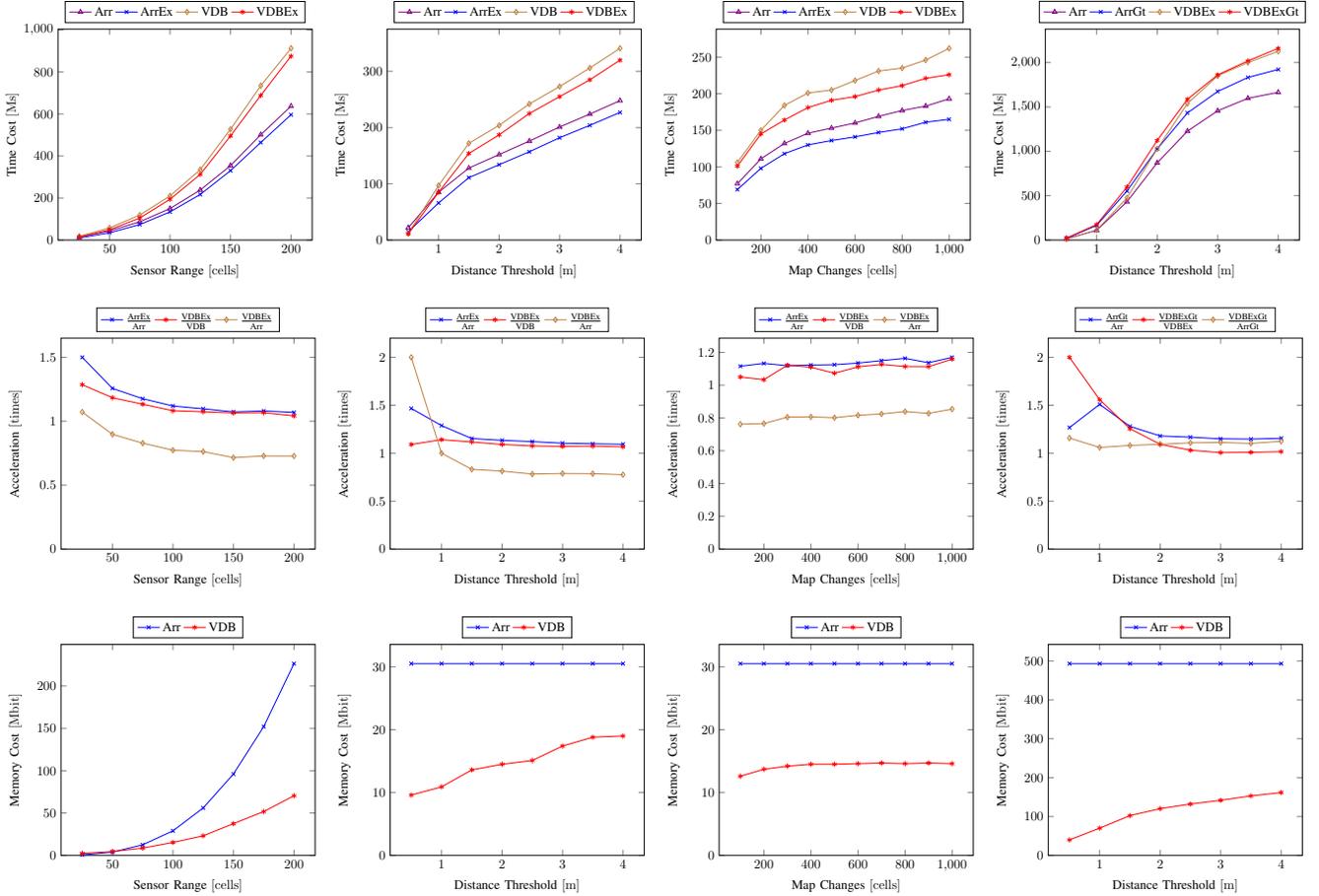
\begin{figure*}[!ht]
	\centering
	\subfloat{
		\begin{tikzpicture}[scale=0.5]
		\pgfplotsset{every axis legend/.append style={
				at={(0.5,1.03)},
				anchor=south}}	
		
		\begin{axis}[use units,
		x unit=cells,
		y unit=s,y unit prefix=M,
		xlabel=Sensor Range,
		ylabel=Time Cost,
		legend columns=6,
		ymin = 0]
		\addplot[color=violet,mark=triangle] coordinates {
			(25,15) (50,44) (75,87) (100,150) (125,238) (150,354) (175,501) (200, 637)
		};
		\addplot[color=blue,mark=x] coordinates {
			(25,10) (50,35) (75,74) (100,134) (125,217) (150,330) (175,464) (200, 596)
		};
		\addplot[color=brown,mark=diamond] coordinates {
			(25,18) (50,58) (75,119) (100,210) (125,335) (150,527) (175,734) (200, 912)
		};
		\addplot[color=red,mark=asterisk] coordinates {
			(25,14) (50,49) (75,105) (100,194) (125,312) (150,495) (175,688) (200, 875)
		};
		\legend{Arr, ArrEx, VDB, VDBEx}
		\end{axis}			
		\end{tikzpicture}
	}
	\subfloat{
		\begin{tikzpicture}[scale=0.5]
		\pgfplotsset{every axis legend/.append style={
				at={(0.5,1.03)},
				anchor=south}}					
		\begin{axis}[use units,
		x unit=m,
		y unit=s,y unit prefix=M,
		xlabel=Distance Threshold,
		ylabel=Time Cost,
		legend columns=6, ymin = 0]
		\addplot[color=violet,mark=triangle] coordinates {
			(0.5,22) (1.0,85) (1.5,128) (2.0,152) (2.5,176) (3.0,201) (3.5,224) (4.0, 248)
		};
		\addplot[color=blue,mark=x] coordinates {
			(0.5,15) (1.0,66) (1.5,111) (2.0,134) (2.5,157) (3.0,182) (3.5,204) (4.0, 227)
		};
		\addplot[color=brown,mark=diamond] coordinates {
			(0.5,12) (1.0,97) (1.5,172) (2.0,204) (2.5,242) (3.0,273) (3.5,306) (4.0, 341)
		};
		\addplot[color=red,mark=asterisk] coordinates {
			(0.5,11) (1.0,85) (1.5,154) (2.0,187) (2.5,225) (3.0,255) (3.5,285) (4.0, 320)
		};
		\legend{Arr, ArrEx, VDB, VDBEx}
		\end{axis}
		\end{tikzpicture}	
	}
	\subfloat{
		\begin{tikzpicture}[scale=0.5]
		\pgfplotsset{every axis legend/.append style={
				at={(0.5,1.03)},
				anchor=south}}					
		\begin{axis}[use units,
		x unit=cells,
		y unit=s,y unit prefix=M,
		xlabel=Map Changes,
		ylabel=Time Cost,
		legend columns=6, ymin = 0]
		\addplot[color=violet,mark=triangle] coordinates {
			(100,77) (200,111) (300,132) (400,146) (500,153) (600,160) (700,169) (800, 177) (900,183) (1000,193)
		};
		\addplot[color=blue,mark=x] coordinates {
			(100,69) (200,98) (300,118) (400,130) (500,136) (600,141) (700,147) (800, 152) (900,161) (1000,165)
		};
		\addplot[color=brown,mark=diamond] coordinates {
			(100,106) (200,150) (300,184) (400,201) (500,205) (600,218) (700,231) (800, 235) (900,246) (1000,262)
		};
		\addplot[color=red,mark=asterisk] coordinates {
			(100,101) (200,145) (300,164) (400,181) (500,191) (600,196) (700,205) (800, 211) (900,221) (1000,226)
		};
		\legend{Arr, ArrEx, VDB, VDBEx}
		\end{axis}
		\end{tikzpicture}	
	}
	\subfloat{
	\begin{tikzpicture}[scale=0.5]
		\pgfplotsset{every axis legend/.append style={
				at={(0.5,1.03)},
				anchor=south}}					
		\begin{axis}[use units,
		x unit=m,
		y unit=s,y unit prefix=M,
		xlabel=Distance Threshold,
		ylabel=Time Cost,
		legend columns=6, ymin = 0]
		\addplot[color=violet,mark=triangle] coordinates {
			(0.5,15) (1.0,110) (1.5,432) (2.0,870) (2.5,1227) (3.0,1456) (3.5,1597) (4.0, 1664)
		};
		\addplot[color=blue,mark=x] coordinates {
			(0.5,19) (1.0,166) (1.5,553) (2.0,1027) (2.5,1432) (3.0,1674) (3.5,1832) (4.0, 1922)
		};
		\addplot[color=brown,mark=diamond] coordinates {
			(0.5,11) (1.0,113) (1.5,476) (2.0,1027) (2.5,1538) (3.0,1850) (3.5,2001) (4.0, 2127)
		};
		\addplot[color=red,mark=asterisk] coordinates {
			(0.5,22) (1.0,176) (1.5,598) (2.0,1122) (2.5,1586) (3.0,1861) (3.5,2019) (4.0, 2160)
		};
		\legend{Arr, ArrGt, VDBEx, VDBExGt}
		\end{axis}
		\end{tikzpicture}	
	}

	\subfloat{
		\begin{tikzpicture}[scale=0.5]
		\pgfplotsset{every axis legend/.append style={
				at={(0.5,1.03)},
				anchor=south}}	
		
		\begin{axis}[use units,
		x unit=cells,
		y unit=times,
		xlabel=Sensor Range,
		ylabel=Acceleration,
		legend columns=6,
		ymin = 0]
		\addplot[color=blue,mark=x] coordinates {
			(25,1.5) (50,1.257) (75,1.176) (100,1.119) (125,1.097) (150,1.073) (175,1.080) (200, 1.069)
		};
		\addplot[color=red,mark=asterisk] coordinates {
			(25,1.286) (50,1.184) (75,1.133) (100,1.082) (125,1.074) (150,1.065) (175,1.067) (200, 1.042)
		};
		\addplot[color=brown,mark=diamond] coordinates {
			(25,1/0.933) (50,1/1.114) (75,1/1.207) (100,1/1.293) (125,1/1.312) (150,1/1.398) (175,1/1.373) (200, 1/1.374)
		};

		\legend{$\frac{\text{ArrEx}}{\text{Arr}}$, $\frac{\text{VDBEx}}{\text{VDB}}$, $\frac{\text{VDBEx}}{\text{Arr}}$}
		\end{axis}			
		\end{tikzpicture}
	}
	\subfloat{
		\begin{tikzpicture}[scale=0.5]
		\pgfplotsset{every axis legend/.append style={
				at={(0.5,1.03)},
				anchor=south}}					
		\begin{axis}[use units,
		x unit=m,
		y unit=times,
		xlabel=Distance Threshold,
		ylabel=Acceleration,
		legend columns=6, ymin = 0]
		\addplot[color=blue,mark=x] coordinates {
			(0.5,1.467) (1.0,1.288) (1.5,1.153) (2.0,1.134) (2.5,1.121) (3.0,1.104) (3.5,1.098) (4.0, 1.093)
		};
		\addplot[color=red,mark=asterisk] coordinates {
			(0.5,1.091) (1.0,1.141) (1.5,1.117) (2.0,1.091) (2.5,1.076) (3.0,1.071) (3.5,1.074) (4.0, 1.066)
		};
		\addplot[color=brown,mark=diamond] coordinates {
			(0.5,1/0.5) (1.0,1/1.0) (1.5,1/1.203) (2.0,1/1.230) (2.5,1/1.278) (3.0,1/1.269) (3.5,1/1.272) (4.0, 1/1.290)
		};
		\legend{$\frac{\text{ArrEx}}{\text{Arr}}$, $\frac{\text{VDBEx}}{\text{VDB}}$, $\frac{\text{VDBEx}}{\text{Arr}}$}
		\end{axis}
		\end{tikzpicture}	
	}
	\subfloat{
		\begin{tikzpicture}[scale=0.5]
		\pgfplotsset{every axis legend/.append style={
				at={(0.5,1.03)},
				anchor=south}}					
		\begin{axis}[use units,
		x unit=cells,
		y unit=times,
		xlabel=Map Changes,
		ylabel=Acceleration,
		legend columns=6, ymin = 0]
		\addplot[color=blue,mark=x] coordinates {
			(100,1.116) (200,1.133) (300,1.119) (400,1.123) (500,1.125) (600,1.135) (700,1.150) (800, 1.164) (900,1.137) (1000,1.170)
		};
		\addplot[color=red,mark=asterisk] coordinates {
			(100,1.050) (200,1.034) (300,1.122) (400,1.110) (500,1.073) (600,1.112) (700,1.127) (800, 1.114) (900,1.113) (1000,1.159)
		};
		\addplot[color=brown,mark=diamond] coordinates {
			(100,1/1.312) (200,1/1.306) (300,1/1.242) (400,1/1.240) (500,1/1.248) (600,1/1.225) (700,1/1.213) (800, 1/1.192) (900,1/1.208) (1000,1/1.171)
		};
		\legend{$\frac{\text{ArrEx}}{\text{Arr}}$, $\frac{\text{VDBEx}}{\text{VDB}}$, $\frac{\text{VDBEx}}{\text{Arr}}$}
		\end{axis}
		\end{tikzpicture}	
	}
	\subfloat{
		\begin{tikzpicture}[scale=0.5]
		\pgfplotsset{every axis legend/.append style={
				at={(0.5,1.03)},
				anchor=south}}					
		\begin{axis}[use units,
		x unit=m,
		y unit=times,
		xlabel=Distance Threshold,
		ylabel=Acceleration,
		legend columns=6, ymin = 0]
		\addplot[color=blue,mark=x] coordinates {
			(0.5,1.267) (1.0,1.509) (1.5,1.280) (2.0,1.180) (2.5,1.167) (3.0,1.150) (3.5,1.147) (4.0, 1.155)
		};
		\addplot[color=red,mark=asterisk] coordinates {
			(0.5,2.0) (1.0,1.558) (1.5,1.256) (2.0,1.093) (2.5,1.031) (3.0,1.006) (3.5,1.009) (4.0, 1.016)
		};
		\addplot[color=brown,mark=diamond] coordinates {
			(0.5,1.158) (1.0,1.060) (1.5,1.081) (2.0,1.093) (2.5,1.108) (3.0,1.112) (3.5,1.102) (4.0,1.124)
		};
		\legend{
			$\frac{\text{ArrGt}}{\text{Arr}}$,$\frac{\text{VDBExGt}}{\text{VDBEx}}$, $\frac{\text{VDBExGt}}{\text{ArrGt}}$}
		\end{axis}
		\end{tikzpicture}	
	}
	
	\subfloat{
		\begin{tikzpicture}[scale=0.5]
		\pgfplotsset{every axis legend/.append style={
				at={(0.5,1.03)},
				anchor=south}}					
		\begin{axis}[use units,
		x unit=cells,
		y unit=bit,y unit prefix=M,
		xlabel=Sensor Range,
		ylabel=Memory Cost,
		legend columns=6, ymin = 0]
		\addplot[color=blue,mark=x] coordinates {
			(25,0.62) (50,3.9) (75,12.5) (100,29.0) (125,56.0) (150,96.1) (175,152.0) (200, 226.3)
		};
		\addplot[color=red,mark=asterisk] coordinates {
			(25,2.4) (50,4.6) (75,8.6) (100,15.3) (125,23.1) (150,37.5) (175,51.7) (200, 70.4)
		};
		\legend{Arr, VDB}
		\end{axis}
		\end{tikzpicture}	
	}
	\subfloat{
		\begin{tikzpicture}[scale=0.5]
		\pgfplotsset{every axis legend/.append style={
				at={(0.5,1.03)},
				anchor=south}}					
		\begin{axis}[use units,
		x unit=m,
		y unit=bit,y unit prefix=M,
		xlabel=Distance Threshold,
		ylabel=Memory Cost,
		legend columns=6, ymin = 0]
		\addplot[color=blue,mark=x] coordinates {
			(0.5,30.5) (1.0,30.5) (1.5,30.5) (2.0,30.5) (2.5,30.5) (3.0,30.5) (3.5,30.5) (4.0, 30.5)
		};
		\addplot[color=red,mark=asterisk] coordinates {
			(0.5,9.6) (1.0,10.9) (1.5,13.6) (2.0,14.5) (2.5,15.1) (3.0,17.4) (3.5,18.8) (4.0, 19.0)
		};
		\legend{Arr, VDB}
		\end{axis}
		\end{tikzpicture}	
	}
	\subfloat{
		\begin{tikzpicture}[scale=0.5]
		\pgfplotsset{every axis legend/.append style={
				at={(0.5,1.03)},
				anchor=south}}					
		\begin{axis}[use units,
		x unit=cells,
		y unit=bit,y unit prefix=M,
		xlabel=Map Changes,
		ylabel=Memory Cost,
		legend columns=6, ymin = 0]
		\addplot[color=blue,mark=x] coordinates {
			(100,30.5) (200,30.5) (300,30.5) (400,30.5) (500,30.5) (600,30.5) (700,30.5) (800, 30.5) (900,30.5) (1000,30.5)
		};
		\addplot[color=red,mark=asterisk] coordinates {
			(100,12.6) (200,13.7) (300,14.2) (400,14.5) (500,14.5) (600,14.6) (700,14.7) (800, 14.6) (900,14.7) (1000,14.6)
		};
		\legend{Arr, VDB}
		\end{axis}
		\end{tikzpicture}	
	}
	\subfloat{
		\begin{tikzpicture}[scale=0.5]
		\pgfplotsset{every axis legend/.append style={
				at={(0.5,1.03)},
				anchor=south}}					
		\begin{axis}[use units,
		x unit=m,
		y unit=bit,y unit prefix=M,
		xlabel=Distance Threshold,
		ylabel=Memory Cost,
		legend columns=6, ymin = 0]
		\addplot[color=blue,mark=x] coordinates {
			(0.5,493) (1.0,493) (1.5,493) (2.0,493) (2.5,493) (3.0,493) (3.5,493) (4.0, 493)
		};
		\addplot[color=red,mark=asterisk] coordinates {
			(0.5,39.8) (1.0,70.1) (1.5,102.3) (2.0,120.2) (2.5,132.1) (3.0,141.7) (3.5,153) (4.0, 161.6)
		};
		\legend{Arr, VDB}
		\end{axis}
		\end{tikzpicture}	
	}		
	\caption{Performance of different EDT implementations. The prefix \texttt{Arr-} and \texttt{VDB-} indicates the underlying data structures. The \texttt{-Ex} suffix indicates that the transform procedures in Alg. \ref{alg} are adopted. The \texttt{-Gt} suffix indicates a global transform is adopted. Implementations with the same data structure have the same memory cost, thus only the \texttt{Arr} and \texttt{VDB} are reported in the third row. } 
	\label{controlexp}			
\end{figure*}

\section{Experiments}
\label{secExp}
In this section, we first conduct some ablation studies to investigate the performance of VDB-EDT under different transform settings, and then make an overall comparison with the state-of-the-art method. Lastly, we demonstrate an example application of online trajectory generation based on VDB-EDT. All the experiments are conducted with a single computing thread on the i7-5930K (3.50GHz) CPU.
\subsection{Ablation Study}
\label{simexp}
The performance of EDT algorithm is primarily related to three factors: {sensor ranges}, {distance thresholds}, and {map changes} (the number of dynamically changed map cells). We thus design three control experiments to quantify their influence. To precisely control the changes of these factors, a simulated cubic map is leveraged in the experiments, and the map resolution is fixed to 0.2m/cell. Each experiment is repeated 40 times and the average performance is presented in the first three columns of Fig. \ref{controlexp}.
A commonly-used general EDT \cite{lau2010improved} (denoted without \texttt{-Ex} suffix) is taken as the baseline to evaluate the runtime efficiency of the proposed algorithm (denoted with \texttt{-Ex} suffix). For each algorithm, we also provide two implementations based on the array and VDB data structures to compare their memory efficiency (denoted with \texttt{Arr-} and \texttt{VDB-} prefix, respectively). 

\textbf{Sensor Ranges}: The range value varies from 25 to 200 cells, representing a real distance (side length of the cubic map) of 5m to 40m. Here, the cubic map simulates a local map within the sensor range, thus a large range value indicates a large local map. The distance threshold is fixed to 10 cells (2m). In terms of map changes, we first build a distance field based on the cubic map with 500 randomly placed obstacles (cells). We then randomly remove half of the obstacles and add an equal number of new ones to simulate the dynamic changes in the map. Lastly, the distance field is incrementally updated, and the results are presented in the first column of Fig. \ref{controlexp}.  

\textbf{Distance Thresholds}: Eight threshold values ranging from 3 cells (0.6m) to 20 cells (4.0m) are tested. The sensor range is fixed to 100 cells, which yields a cubic map with a side length of 20m. The same number of obstacles and dynamic changes as the first experiment are leveraged to build the distance field. The results are presented in the second column of Fig. \ref{controlexp}.

\textbf{Map Changes}: The sensor range is fixed to 100 cells (20m) and the distance threshold is fixed to 10 cells (2m). The cubic maps are built with a different number of obstacles, varying from 100 to 1000 cells, and the same strategy for simulating map changes as the first experiment is used to built distance fields. The result is presented in the third column of Fig. \ref{controlexp}.

The three rows in Fig. \ref{controlexp} show the time cost, relative acceleration of running speed, and the memory cost of the EDTs, respectively. By comparing the \texttt{-Ex} algorithm (ours) with \texttt{non-Ex} one \cite{lau2010improved}, we can see that our proposed method always presents a better runtime efficiency, and the running speed is accelerated by about 1.1x-1.5x (see the red and blue lines in the second row), which demonstrates the effectiveness of our proposed transform strategies. Meanwhile, it is shown that the acceleration rate tends to decrease when the sensor range and distance threshold are increased. The reasons are as follows: 1) The obstacles scatter more sparsely in large cubic maps (senor ranges), and thus there are fewer opportunities that allow the proposed method to take effect, namely the lowering waves stop the raising waves at an early stage; 2) When the distance threshold is increased, more transform steps are needed for the waves to arrive at wave boundaries, which means most computing time is spent on processing the large quantity of non-boundary cells, whereas our method takes effect mainly at wave boundaries (see the first case in the $Lower$ function), thus the improvement cannot be well highlighted after dividing by the large cost of non-boundary cells. Contrariwise, the acceleration rate ascends when map changes are enlarged. The reason is that the increment of map changes provides more chances for the lowering waves to stop the raising waves at an early stage, thus avoiding more unnecessary computations. Through comparing the \texttt{VdbEx} (VDB-EDT) with \texttt{Arr} (array-based conventional EDT), we can see that the running speed of VDB-EDT decreases by about 10$\%$-25$\%$ (see the brown lines in the second row), while the memory cost is reduced by about 30$\%$-60$\%$ (see the third row). Herein, the increment of time cost is inevitable, as VDB is based on tree structures and has a slower random access speed than the array-based implementation. Theoretically, the memory cost of array-based EDT provides an upper bound for that of VDB-EDT. In a fully dense environment with a full distance transform, the VDB becomes a fully allocated tree, which will lead to the same memory cost as the array.
\begin{table*}[t]
	\setlength\arrayrulewidth{0.7pt}
	\renewcommand{\arraystretch}{1.3}
	\newcolumntype{M}{>{\centering\arraybackslash}m{\dimexpr.051\linewidth-0.8\tabcolsep}}
	\newcolumntype{X}{>{\centering\arraybackslash}m{\dimexpr.090\linewidth-0.8\tabcolsep}}
	\newcolumntype{N}{>{\centering\arraybackslash}m{\dimexpr.080\linewidth-0.8\tabcolsep}}
	\newcolumntype{S}{>{\centering\arraybackslash}m{\dimexpr.035\linewidth-0.8\tabcolsep}}
	\centering
	\caption{Performance on OctoMap dataset \cite{freiburg}. 'Obst.' column records the number of obstacles in the map and 'Res.' indicates the map resolutions. The time and memory costs are measured by \textit{Millisecond (Ms)} and \textit{Megabyte (Mb)}, respectively.}
	\label{table:result}
	\begin{tabular}{X|NNS|MMMM|MMMM|MM}
		\hline
		\textbf{Dataset}& \multicolumn{3}{c|}{\textbf{Map Information}} &\multicolumn{4}{c|}{\textbf{Global Transform} \textit{(Ms)}}& \multicolumn{4}{c|}{\textbf{Incremental Transform} \textit{(Ms)}} & \multicolumn{2}{c}{\textbf{Memory} \textit{(Mb)}}  \\
		\hline
		Name& Size & Obst.  & Res. & Arr & ArrEx & VDB & VDBEx & Arr & ArrEx & VDB & VDBEx & Arr & VDBEx \\
		\hline
		fr-078      & 256x282x95	& 287664   & 0.05 & 814  & 838	 & 940	 & 954		& 863	& 819	& 942  	& 898  &209.3	&148.6	\\
		fr-079      & 875x364x66 	& 401373   & 0.05 & 1893 & 1935	 & 2075	 & 2093 	& 1179 	& 1117  & 1355 	& 1278 &641.5	&321.8	\\
		fr-campus   & 1465x837x140 	& 1198329  & 0.20 & 4441 & 4621	 & 4240  & 4333 	& 723  	& 664 	& 839 	& 759  &5222.4 	&782.2 \\
		new-college & 808x1250x165  & 536458   & 0.20 & 2175 & 2246	 & 2057  & 2113		& 461  	& 418 	& 589 	& 517  &5120.0 	&471.3	\\																
		\hline
	\end{tabular}
\end{table*} 
\begin{figure}[t]
	\centering
		\captionsetup{belowskip=-10pt}
	\includegraphics[width=0.38\textwidth]{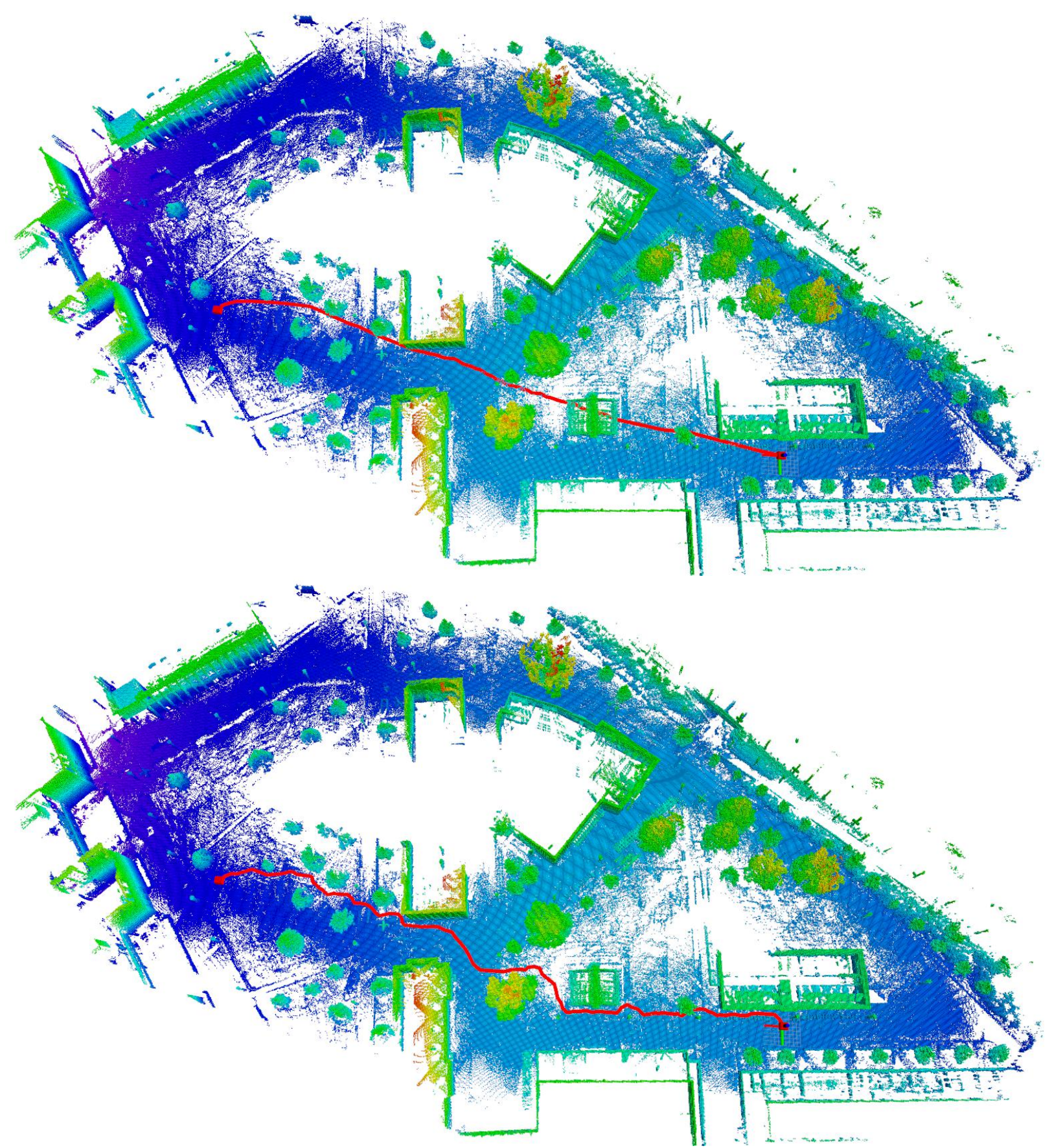}
	\caption{Demonstration of long-range pathfinding based on the global transform on a static map in \cite{freiburg}. In the top figure, only the path length is considered, while in the bottom figure, a clearance cost is additionally integrated to improve the safety.}
	\label{archt}
\end{figure}

\textbf{Global Transform v.s. Incremental Transform}: The above experiments are based on incremental transform, in which the \textit{lower} and \textit{raise} functions iteratively execute to update the distance field. In the global transform, the entire distance field is recomputed by leveraging the \textit{lower} function. Since the \textit{lower} function is with higher efficiency than the \textit{raise} function, trade-offs exist between choosing global and incremental transforms, which are mainly affected by the distance threshold and the ratio of map changes (the number of removed obstacles divided by that of the added ones). 
According to our experimental studies on a cubic map with a side length of 250 cells, the global transform always exhibits a better time performance when the ratio is larger than 0.25; otherwise, the performance is primarily determined by the distance threshold. We thus design the fourth experiment to quantify the performance of different transform types, in which a distance field is first built based on the cubic map with 800 randomly placed obstacles, and then we test the performance of globally and incrementally updating the distance field after removing 160 obstacles and adding 400 new obstacles in the map. 
The experiment results are presented in the last column of Fig. \ref{controlexp}. 
It can be seen that the advantages of incremental transforms (denoted without \texttt{-Gt} suffix) gradually disappear as the distance threshold increases (see the second row). This is due to the fact that large threshold values typically require more transform steps, which leads to a larger overall time cost of the \textit{raise} function than that of the \textit{lower} function, making the incremental transform lost its advantages. 
Compared with the VDB-EDT (the red line in the second row), the array-based conventional EDT (the blue in the second row) shows a smaller decreasing rate, which indicates a less efficient design of the \textit{lower} function in \cite{lau2010improved}, and this can also be verified by a direct comparison between the \textit{lower} functions (the brown lines in the second row).
The fourth experiment provides a reference for the selection of global or incremental transforms under different application scenarios.

In addition to the experiments on simulated cubic maps, we conduct another experiment on real-world maps in the OctoMap dataset \cite{freiburg} to further evaluate the performance of VDB-EDT. In this experiment, the distance threshold is set to 2m, and both the global and incremental transforms are evaluated. When testing the incremental transform, we randomly remove 10,000 obstacles and then add an equal number of new ones to update the distance field. The experiment results are presented in Table \ref{table:result}, which conveys the same information as the previous experiments: 1) The proposed algorithm can effectively facilitate the distance transform procedure and accelerate the running speed of conventional EDTs; 2) The VDB data structure can significantly reduce the memory consumption of EDTs; 3) The VDB-EDT exhibits competitive time performance with the array-based EDT while achieving significant improvement in memory efficiency, 
especially in the \textit{fr-campus} and \textit{new-college} maps.
 

\begin{table*}[t]
	\setlength\arrayrulewidth{0.7pt}
	\renewcommand{\arraystretch}{1.3}
	\newcolumntype{M}{>{\centering\arraybackslash}m{\dimexpr.062\linewidth-0.8\tabcolsep}}
	\newcolumntype{X}{>{\centering\arraybackslash}m{\dimexpr.097\linewidth-0.8\tabcolsep}}
	\newcolumntype{N}{>{\centering\arraybackslash}m{\dimexpr.104\linewidth-0.8\tabcolsep}}
	\newcolumntype{E}{>{\centering\arraybackslash}m{\dimexpr.045\linewidth-0.8\tabcolsep}}
	\centering
	\caption{Experiment results on the cow-and-lay and SubT dataset. 'Res.', 'Succ.', and 'Var.' are abbreviations of \textit{resolution}, \textit{success}, and \textit{variance}, respectively. The runtime and memory cost are measured with \textit{Millisecond (Ms)} and \textit{Megabyte (Mb)}, respectively.}
	\begin{tabular}{X|MEM|MMM|MMMM|N}
		\hline
		\textbf{Dataset}& \multicolumn{3}{c|}{\textbf{Basic Information}} &\multicolumn{3}{c|}{\textbf{Processed Cells}}& \multicolumn{4}{c|}{\textbf{Runtime} \textit{(Ms)}} & \textbf{Memory} \textit{(Mb)} \\   
		\hline
		Name & Method  & Succ. & Frames & Changed & Lowered & Raised   & Mean & Var. & Min & Max\\
		\hline	
		cow-and-lady  & FIESTA   & Y  & 2194   & 526.4   & 29393.2  & 15264.4  & 47.527 & 48.737  	& 0.012 & 256.884 & 309.2 \\
		cow-and-lady  & Vdb	     & Y  & 2065   & 2867.3  & 71528.2	& 5869.9   & 34.237 & 20.987	& 2.285	& 184.588 & 221.3 \\
		cow-and-lady  & VdbEx	 & Y  & 2112   & 2847.3  & 50567.1	& 4253.1   & 23.661 & 12.976 	& 1.599 & 137.243 & 221.3 \\
		\hline
		SubT &       FIESTA      & N  & 2047    & 3568.8  & 78015.4  & 5929.5   & 123.309 & 24.007  & 0.259 & 1107.24 & 7577.6 \\
		SubT &          Vdb      & Y  & 2336    & 5985.0 & 708311.4  & 57179.6  & 293.510 & 61.835  & 12.535 & 897.115 & 4403.2 \\
		SubT &        VdbEx      & Y  & 2346    & 5966.7 & 483464.8  & 38980.1  & 198.354 & 38.703  & 12.500 & 390.530 & 4403.2 \\																
		\hline
	\end{tabular}
	\label{table1}
\end{table*}

\subsection{Performance on Benchmark Datasets}
In this experiment, we leverage the commonly-used cow-and-lady dataset and our released SubT dataset to compare VDB-EDT with the state-of-the-art method, FIESTA \cite{FIESTA}. 
The SubT dataset is captured from an unstructured tunnel environment (approximately 200m by 150m) provided by the virtual track\footnote{https://github.com/microsoft/AirSim/releases/tag/v1.2.0Linux} of DARPA subterranean challenge\footnote{https://www.darpa.mil/program/darpa-subterranean-challenge}, and it is more suitable for large-scale performance evaluation than the cow-and-lady dataset (about 10m by 10m). Different from the OctoMap dataset, these two datasets are composed of LIDAR range scans stamped with robot poses, and thus a mapping process is needed to integrate the data. Therefore, an online VDB-mapping module is designed to help build occupancy grid maps for the datasets, and the distance maps are constructed along with the mapping process. 

We follow the same experiment settings as FIESTA. The map resolution and sensor range are set to 0.05m/cell and 5m, respectively. The distance field is updated every 0.5 seconds, and a full distance transform, i.e., $d_{max}=\infty$, is performed.  
The detailed experiment results are presented in Table \ref{table1}. The \textit{Frames} column lists how many frames of the range scans are processed by the algorithms. The \textit{Changed}, \textit{Lowered}, and \textit{Raised} columns show an average number of the cells that are dynamically changed, visited by lowering waves, and visited by raising waves in each frame, respectively.

\begin{figure}[t]
	\centering
	\subfloat[VDB-EDT]{
		\includegraphics[width=0.46\linewidth]{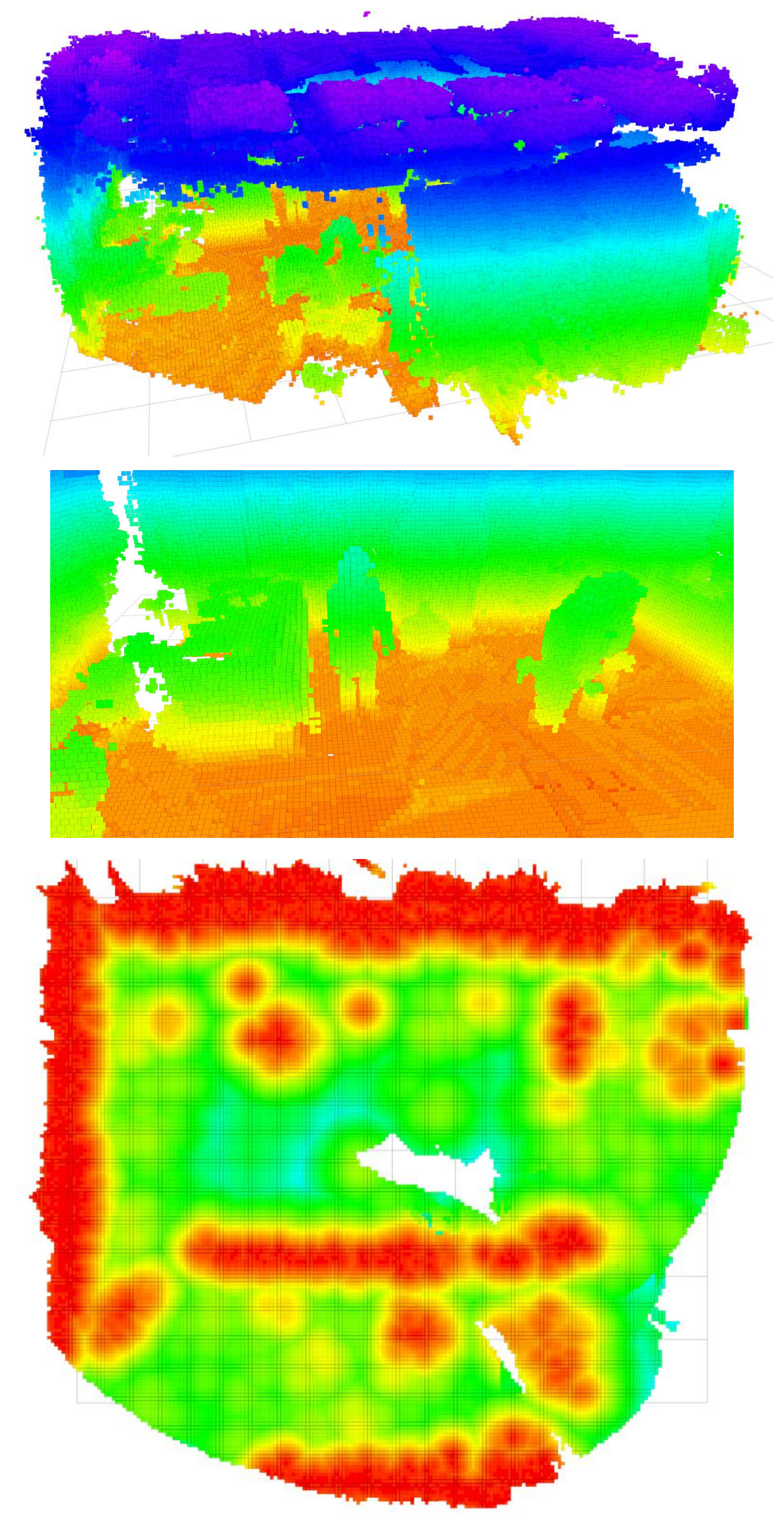}
	}
	\subfloat[FIESTA]{
		\includegraphics[width=0.46\linewidth]{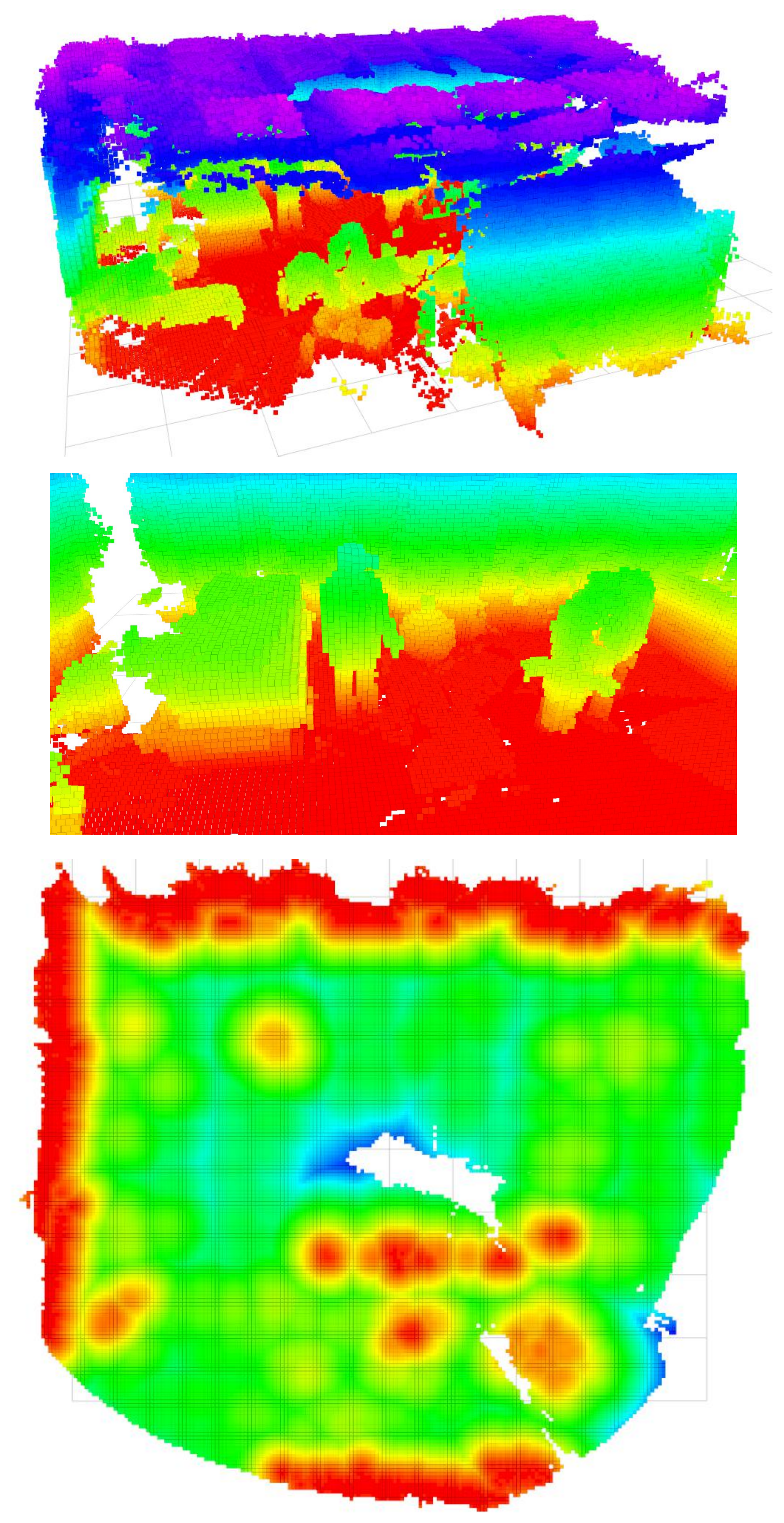}
	}
	\caption{The occupancy grid map (top two rows) and a slice of distance field (bottom row) built from the cow-and-lady dataset. The red color in the distance field indicates a close distance to obstacles.}
	\label{cowlady}
\end{figure}
\begin{figure}[t]
	\captionsetup{belowskip=-10pt}
	\centering
	\subfloat[VDB-EDT]{
		\includegraphics[width=0.46\linewidth]{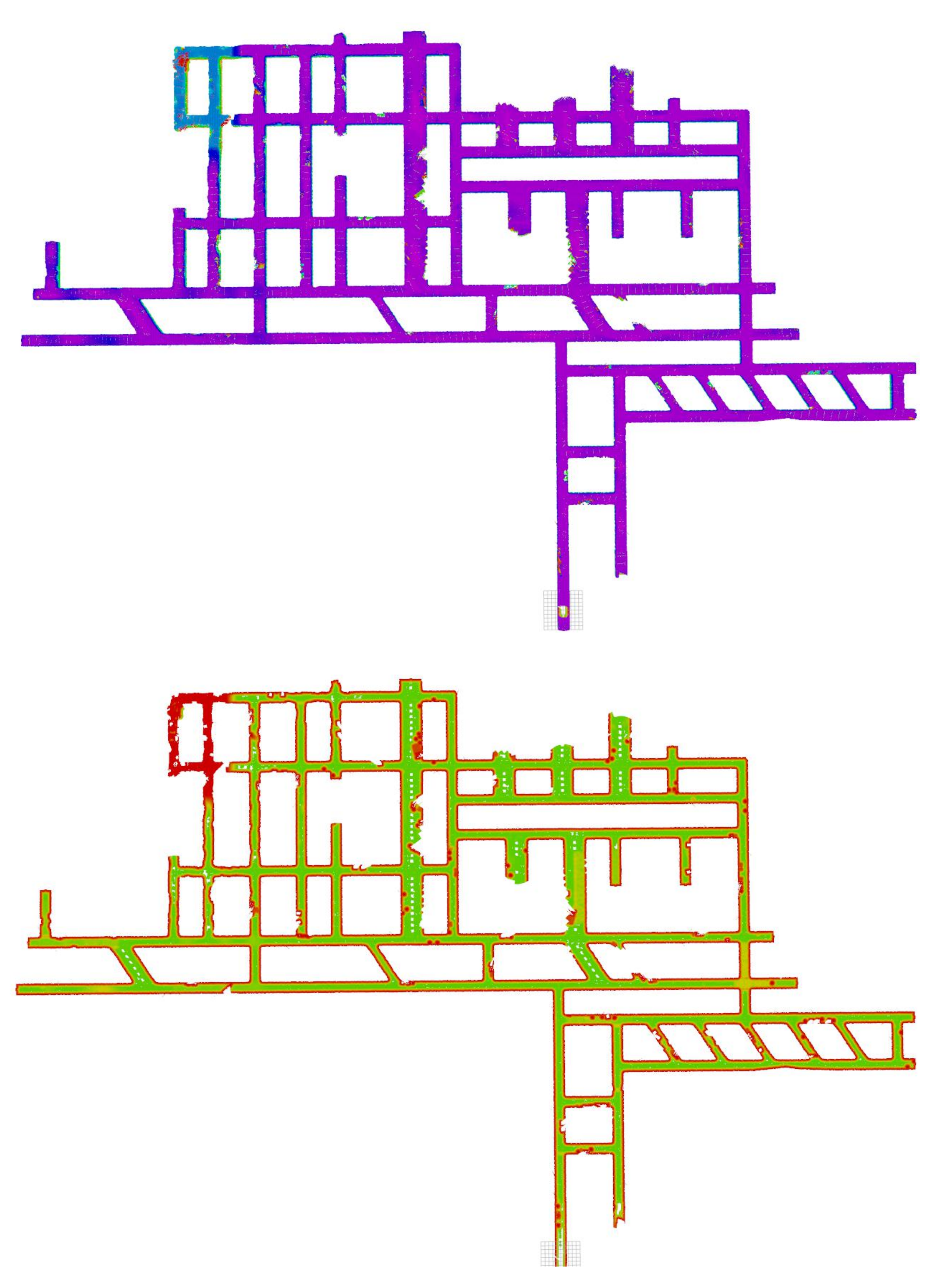}
	}
	\subfloat[FIESTA]{
		\label{subtmap}
		\includegraphics[width=0.46\linewidth]{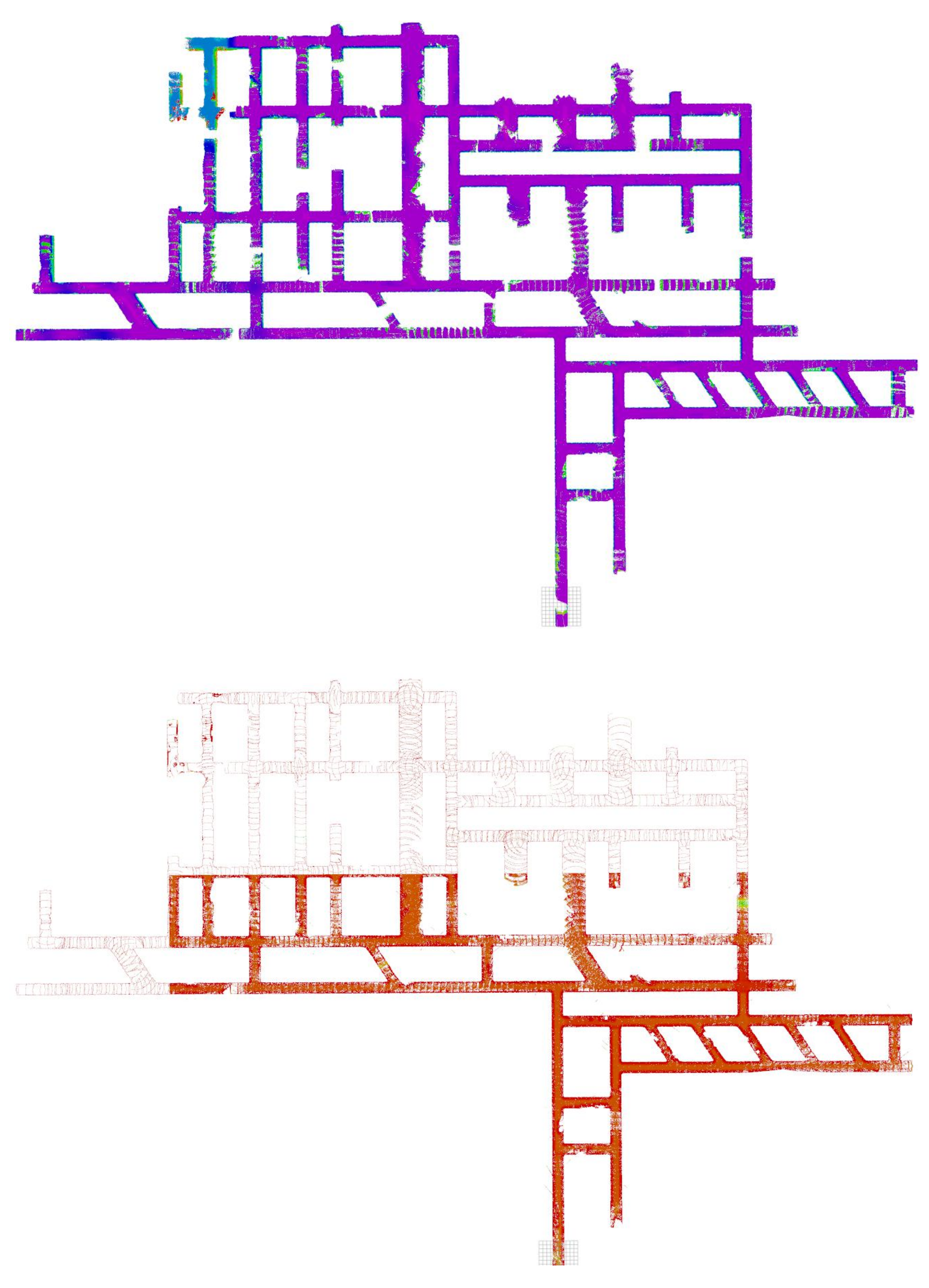}
	}
	\caption{The occupancy grid map (top row) and a slice of distance map (bottom row) built from the SubT dataset.}
	\label{subt}
\end{figure}
\begin{figure*}[t]
	\captionsetup{belowskip=-10pt}
	\centering
	\includegraphics[width=0.95\linewidth]{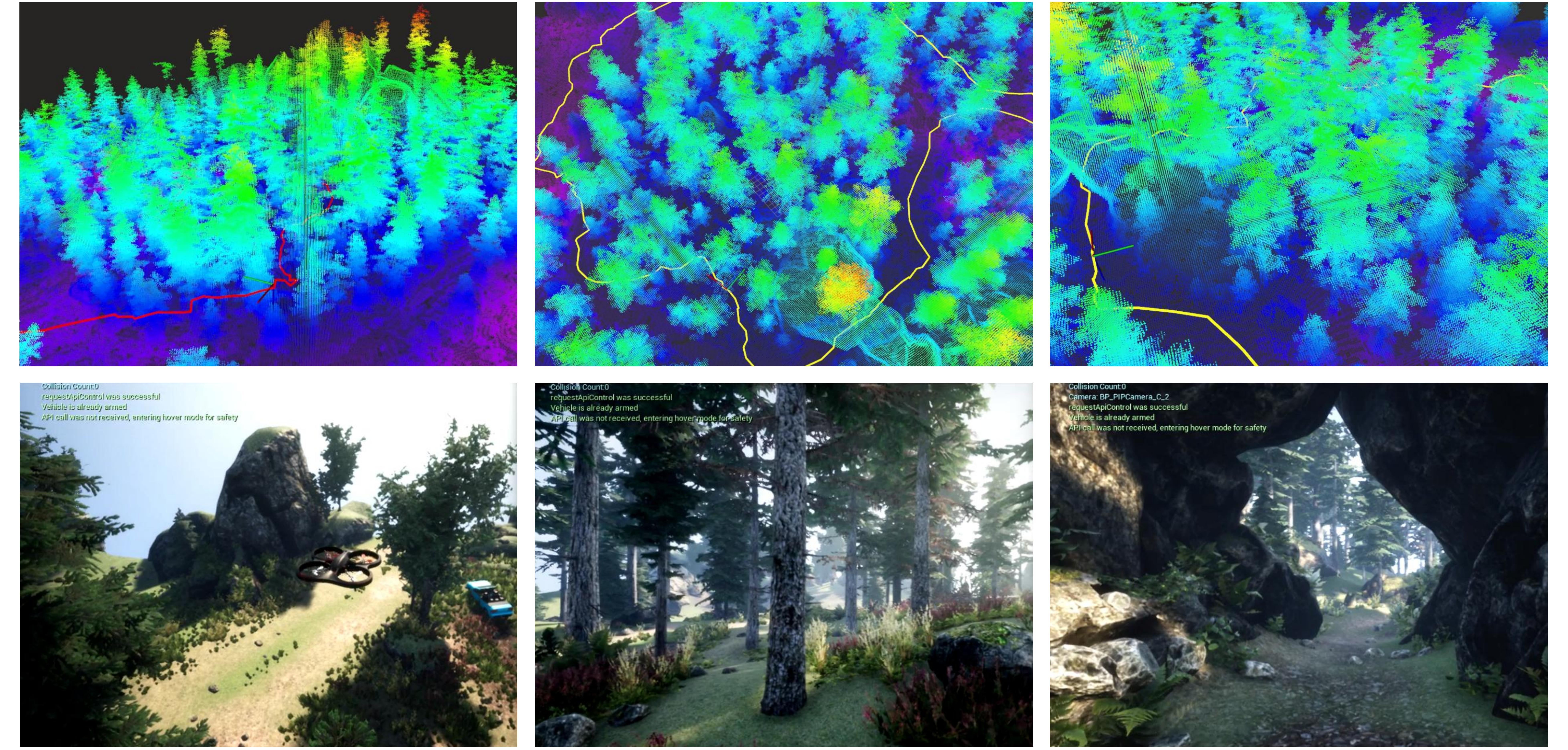}
	\caption{Online planning demonstrations. The first row presents two compound paths denoted with red and yellow colors, respectively. The third image presents a local path segment of the yellow path to demonstrate its clearance. The second row shows a third-person-view image and two first-person-view images of the MAV, the locations of which are denoted by a 3D axis in images of the first row. }
	\label{navi}
\end{figure*} 
As we can see from Table \ref{table1}, VDB-EDT (\texttt{VdbEx}) exhibits a superior performance than FIESTA in both runtime and memory efficiency. Specifically, in the cow-and-lady dataset, VDB-EDT processes 5.4x as many map changes as FIESTA, but only consumes half of its processing time. In the SubT dataset, FIESTA fails to build the correct distance map, but if we only consider the transformed parts, VDB-EDT processes 1.7x more map changes and 1.6x more runtime than FIEST, which again demonstrates the advantages of our method. In terms of memory efficiency, VDB-EDT also shows a significant improvement, benefiting from the efficient hierarchical representation.
To further analyze the results, we visualize the outputs of the algorithms in Fig. \ref{cowlady} and Fig. \ref{subt}. As is shown, the occupancy grid maps built by VDB-EDT are more complete and with higher quality (as more range scans are effectively fused to the maps), which explains why the VDB-EDT processed more map changes. The low-quality mapping results of FIESTA is also the primary reason that leads to the failure of the distance transform (see Fig. \ref{subtmap}). We also notice that the performance of VDB-EDT on the cow-and-lady dataset is much better than that on the SubT dataset. The reasons are as follows: 1) The inaccurate robot poses in cow-and-lady dataset yield an inconsistent alignment between consecutive range scans, and thus make a large number of occupied cells freed, which induces the heavy usage of the $Raise$ function; 2) In FIESTA, once an obstacle is removed, the $Raise$ function will clear all the affected cells (indexed by a doublely linked list) and meanwhile check their neighbors to detect possible wave boundaries, and after that, the $Lower$ function starts to update these cleared cells; 3) Such a process induces much redundancy since a lot of cells are revisited by both of the functions, and moreover, the queries in FIESTA rely on a more complicated process than VDB, resulting in the bad performance of FIESTA in the cow-and-lady dataset.  
 
Through comparing VDB-EDT (\texttt{VdbEx}) with the conventional EDT (\texttt{Vdb}), we can see that VDB-EDT achieves about 1.5x faster running speed (see \textit{Runtime} column), and the number of query operations is reduced by 1/3 (see \textit{Lowered} and \textit{Raised} columns), which significantly demonstrates the effectiveness of our proposed method. Notice that the performance here is much better than that on the simulation dataset in Section \ref{simexp}, where the acceleration rate lies between 1.1x and 1.5x. The reason is that map changes in the benchmark datasets are located more closely, which thus provides more chances for our algorithm to take effect. This is consistent with our analysis of the algorithm in the penultimate paragraph in Section \ref{methodalg}.

\subsection{Online Planning Demonstration}
In this experiment, we specify a series of waypoints in a forest environment, and the MAV is required to safely navigate through these waypoints by solving the two optimization problems expressed in Eq. (\ref{edt}) and (\ref{plan}). Each pair of consecutive waypoints spans a map range of 50x50x100 (cells). The map resolution is 0.25m, and the distance threshold is set to 2m. With these settings, we achieve a planning frequency of 16Hz, and the path segments are presented in Fig. \ref{navi}. As is indicated, the MAV successfully navigates through the forest and even the narrow space under a bridge. Although obstacles are densely distributed in the map, this planning pipeline only costs 1.3Gbit memory, which includes both the occupancy grid map and the incrementally built distance map.

\section{Conclusions}
\label{conclusion}
In this work, we present an algorithm called VDB-EDT to address the distance transform problem. The algorithm is implemented based on a memory-efficient data structure and a novel distance transform procedure, which significantly improves the memory and runtime efficiency of EDTs. Extensive experiments on simulation and real-word datasets strongly demonstrate the effectiveness and advantages of our method. This work further pushes the limits of general EDT algorithms and will facilitate the research on VDB-based mapping, distance transform, and safe motion planning.





\bibliographystyle{IEEEtran}
\bibliography{ref}
%



\end{document}